\pdfoutput=1

\documentclass[11pt]{article}

\usepackage[]{ACL2023}
\usepackage{graphicx}

\usepackage{times}
\usepackage{anyfontsize}
\usepackage{latexsym}
\usepackage{amsmath}
\usepackage{amssymb}
\usepackage{multirow}
\usepackage{xcolor}
\usepackage{makecell}
\usepackage{latexsym}
\newcommand{\code}[1]{{\texttt{#1}}}
\usepackage[T1]{fontenc}

\usepackage[utf8]{inputenc}

\usepackage{microtype}

\usepackage{inconsolata}

\title{DORIC : Domain Robust Fine-Tuning for Open Intent Clustering through Dependency Parsing}

\author{Jihyun Lee$^1$, Seungyeon Seo$^1$, Yunsu Kim$^{1,2}$, Gary Geunbae Lee$^{1,2}$ \\
  $^1$Graduate School of Artificial Intelligence, POSTECH, Republic of Korea\\
  $^2$Department of Computer Science and Engineering, POSTECH, Republic of Korea\\
  \texttt{\{jihyunlee, ssy319, yunsu.kim, gblee\}@postech.ac.kr} \\
}

\begin{document}
\maketitle
\begin{abstract}

We present our work on Track 2 in the Dialog System Technology Challenges 11 (DSTC11). DSTC11-Track2 aims to provide a benchmark for zero-shot, cross-domain, intent-set induction. In the absence of in-domain training dataset, robust utterance representation that can be used across domains is necessary to induce users' intentions. To achieve this, we leveraged a multi-domain dialogue dataset to fine-tune the language model and proposed extracting Verb-Object pairs to remove the artifacts of unnecessary information. Furthermore,  we devised the method that generates each cluster's name for the explainability of clustered results. Our approach achieved 3rd place in the precision score and showed superior accuracy and normalized mutual information (NMI) score than the baseline model on various domain datasets. 
\end{abstract}

\section{Introduction}

Understanding the user’s intent plays an important role in task-oriented dialogue (TOD) systems. Traditionally, understanding the user’s intent requires supervised training using intent-annotated dialogue datasets \cite{xu2013understanding, wang2015mining,  kim2017onenet, goo2018slot}. However, for new emerging domains and services, defining the intent set is challenging and also requires an expert’s knowledge. Therefore, finding an automatic method that identifies intents from raw conversational data is desirable to reduce costs.

The intent clustering task of the 11th Dialog System Technology Challenge (DSTC11) aims to provide a realistic benchmark for the intent induction problem. This track evaluates automatic customer intent induction methods from dialogues between human agents and customers, and the DSTC11 challenge participants are required to create a set of intent labels based on the conversations. To provide a realistic setting, the number of intents and domain of the test set are not provided until the development phase ends.

In this paper, we introduce an automatic intent induction framework that effectively utilizes a public TOD dataset. First, we fine-tuned the language model with multi-domain TOD datasets so that it has a domain-robust semantic representation. Here, we extract verbs and object from utterance to remove the artifacts of unnecessary information. For the training, we applied supervised contrastive learning (SCL), which is known to be stable in language model fine-tuning  \cite{gunel2020supervised}. Second, to infer a new intent set from the unseen domain dataset, we applied a clustering technique that groups the utterances based on the fine-tuned embedding model’s representation. Furthermore, we generate a label name for each cluster to obtain an interpretable result. The cluster label generation method could reduce the effort of examining each set manually to understand the clustering results.

In the experiment with the test dataset (finance and banking domain), we achieved 3rd place in terms of precision and demonstrated superior accuracy (ACC) and a higher normalized mutual information (NMI) score than the baseline. Furthermore, the generated clustering labels reasonably explain each cluster. Finally, we analyzed our model with comparable options and demonstrated the result on development, test, and TOD datasets. We named our framework DORIC, which means \textbf{DO}main \textbf{R}obust fine-tuning for open \textbf{I}ntent \textbf{C}lustering through dependency parsing.

\section{Related Work}

\subsection{Intent Classification}
Traditionally, benchmark datasets \cite{price1990evaluation, coucke2018snips, eric2019multiwoz} for intent identification have sufficient labeled datasets for training, and the task has been solved through the classification method. For example, \citet{goo2018slot} classified the intent and slot information using the attention mechanism, and \citet{kim2017onenet} enriched word embeddings by using semantic lexicons and adapted this strategy to the intent classification. In addition, \citet{wang2015mining} grouped the intent of tweets into six categories, used a graph embedding consisting of tweet nodes, and classified their intents. However, labeling the intent for the raw dialogue dataset requires extensive human labor, so building a new labeled intent dataset in the real world is challenging. Therefore, a robust intent induction model that can be applied to a new domain as an unsupervised method is required.

\subsection{Intent induction with unsupervised method}
The representative method of unsupervised intent induction utilizes the clustering method. \citet{liu2021open} is one example of research that enhanced the clustering algorithm. They proposed a balanced score metric to obtain similar-sized clusters in K-means clustering and found proper K-values that were more stable than naive K-means. \citet{chatterjee2020intent} also enhanced the clustering algorithm by utilizing the outlier information of the density-based clustering model, which is called ITER-DBSCAN. Their work shows greater accuracy on imbalanced intent data. 
On the other hand, there has been research that improved the dialogue representations for better clustering results. For example, \citet{perkins2019dialog} iteratively enhanced the dialogue embedding by reflecting the clustering score, and \citet{lin2019deep} proposed a BiLSTM \cite{mesnil2014using} embedding model with margin loss that is effective in detecting unknown intents. However, robust intent induction in diverse domains was not examined in previous research. Therefore, as a strategy for enhancing the embedding dialogue model, we propose the DORIC method, which robustly embeds diverse dialogue domains.
\section{DSTC11 Intent Clustering Task}
In this task, participants are required to assign an intent label to each dialogue turn. A set of dialogues are provided as input, and each turn is pre-labeled with both its speaker role (i.e., Agent or Customer) and dialogue acts (i.e., InformIntent or not). One development dataset and two test datasets are provided, and each dataset consists of approximately 1K customer support spoken conversations with manual transcriptions and annotations. The development dataset derives from an insurance-related customer support service, and each conversation has an average of 70 turns. In addition, the development dataset contains ground truth intent annotations that allow participants to test and evaluate the model. The number of intent types and the domains of the test dataset are not revealed until the development phase ends. Note that no training dataset is given, as this challenge aims to zero-shot intent induction.
\section{Method}

\begin{figure*}
\centering
\label{figure1}
\includegraphics[width=\textwidth]{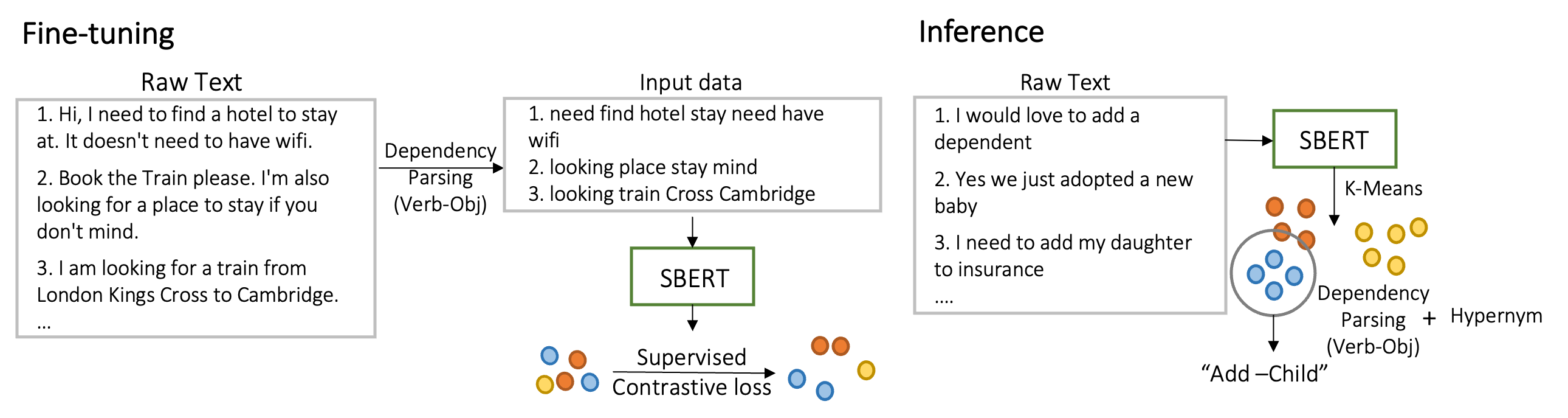}
\caption{Fine-tuning and inference method of our proposed methods. }
\label{fig:method}
\end{figure*}

\subsection{Semantic representation}
With the advance of the pre-trained langue model, leveraging these models for embedding dialogue has exhibited promising results \cite{ham2020end, lin2020mintl, yang2021ubar,lee2022sf}. Following their success, we utilize the pre-trained SBERT \cite{reimers2019sentence} as a backbone model; SBERT has a siamese network structure and performs well in classification \cite{reimers2019sentence}, summarization \cite{zhong2020extractive}, and intent clustering \cite{liu2021open}.

The SBERT model is pre-trained on a written form text dataset that has a different linguistic pattern with the dialogue utterances. This difference could hurt the accuracy on dialogue related tasks \cite{wu2020tod}. Therefore, we fine-tuned the model with the multi-domain public task-oriented dialogue dataset MultiWOZ 2.2 \cite{eric2019multiwoz}. This dataset has nine intents types, and we fine-tuned the model to learn the embedding of utterances according to intent type. In this data, intents are spanned multiple turns, and some utterances contain multiple intents in one utterance. To clarify the match between the utterance and the intent label, we used only the first utterance of the spanned intent dialogue and excluded utterances that contained multiple intents. We analyze the processed fine-tuning dataset in Table~\ref{tab:mwoz}.

\begin{table}[h]
\centering
\small
\begin{tabular}{ll}
\hline
\textbf{Intent} & \textbf{Count} \\
\hline
FindRestaurants   & 3561  \\
SearchHotel   & 3375  \\
FindTrains   & 3262  \\
FindAttractions   & 2795  \\
ReserveHotel   & 1951  \\
GetTrainTickets   & 1926  \\
ReserveRestaurant   & 1600  \\
GetRide   & 1262  \\
FindPolice   & 229  \\
 \hline
Total & 19961   \\
\hline
\end{tabular}
\caption{Type and number of intents of the fine-tuning dataset.}
\label{tab:mwoz}
\end{table}

\begin{table}[h]
\centering
\small
\begin{tabular}{l|ll}
\hline
\textbf{Inference data} & \textbf{NMI}& \textbf{ACC} \\ \hline
Verb-Object & 41.89 & 30.79 \\
Sentence   & \textbf{65.16} & \textbf{56.68} \\
\hline
\end{tabular}
\caption{The comparison of using whole sentence and Verb-Object format in inference. NMI and accuracy result on DSTC11 development are reported.}
\label{tab:pre}
\end{table}

As we aim for unsupervised intent induction, domain robust fine-tuning is crucial to identify the intents across the domain. To do so, we extract Verb-Object structure from the utterance using the dependency parser\footnote{https://spacy.io/}. This additional pre-process removes the effect of non-relevant words or utterance styles when fine-tuning the SBERT model. However, at inference time, we used whole utterances for clustering as preliminary experiments demonstrated better results (Table~\ref{tab:pre}). We demonstrate our overall method in Figure~\ref{fig:method}.

\subsection{Supervised contrastive learning}

\begin{figure}
\centering
\label{figure1}
\begin{tabular}{cc}
\small{Before training} & \small{After training} \\

\includegraphics[width=0.2\textwidth]{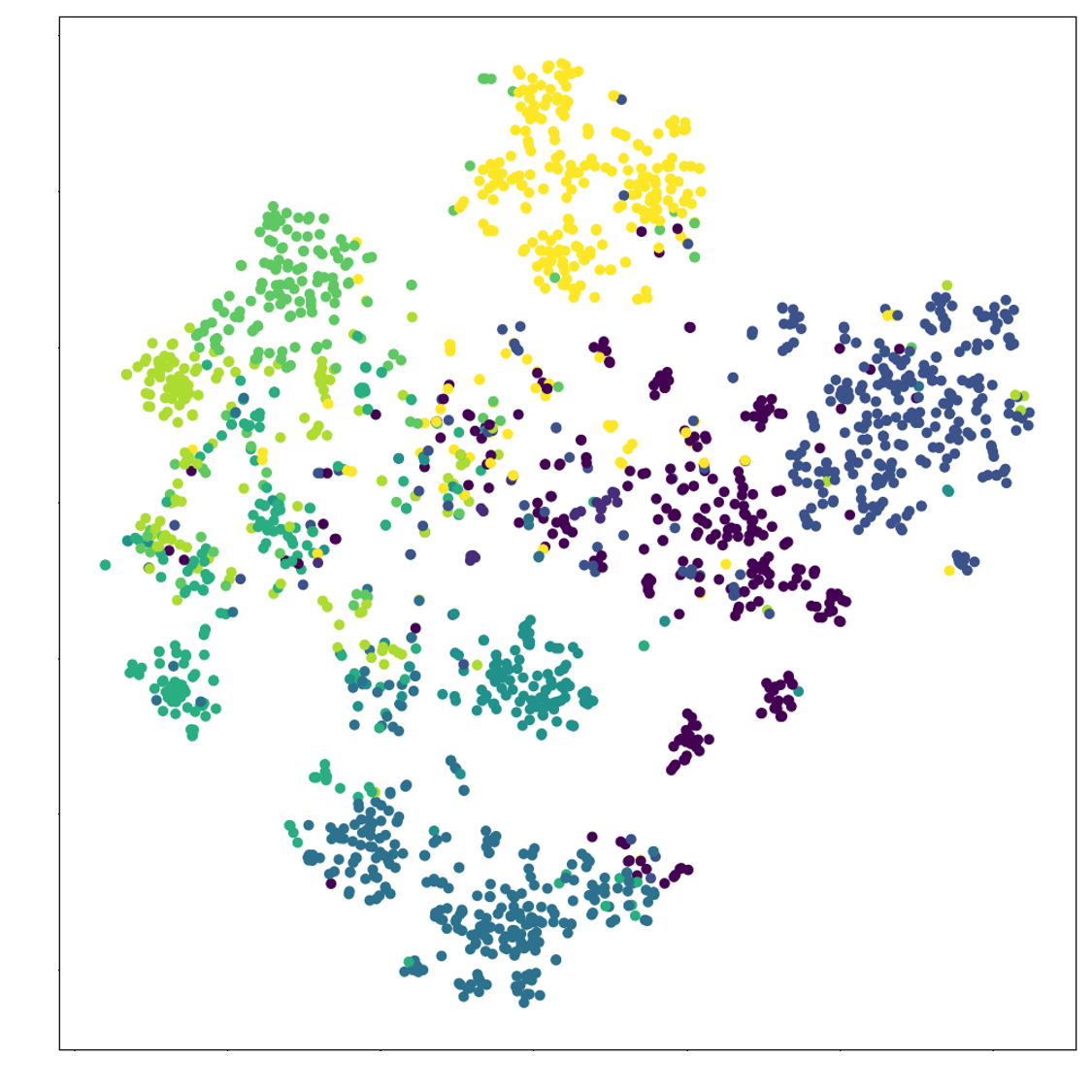}& \includegraphics[width=0.2\textwidth]{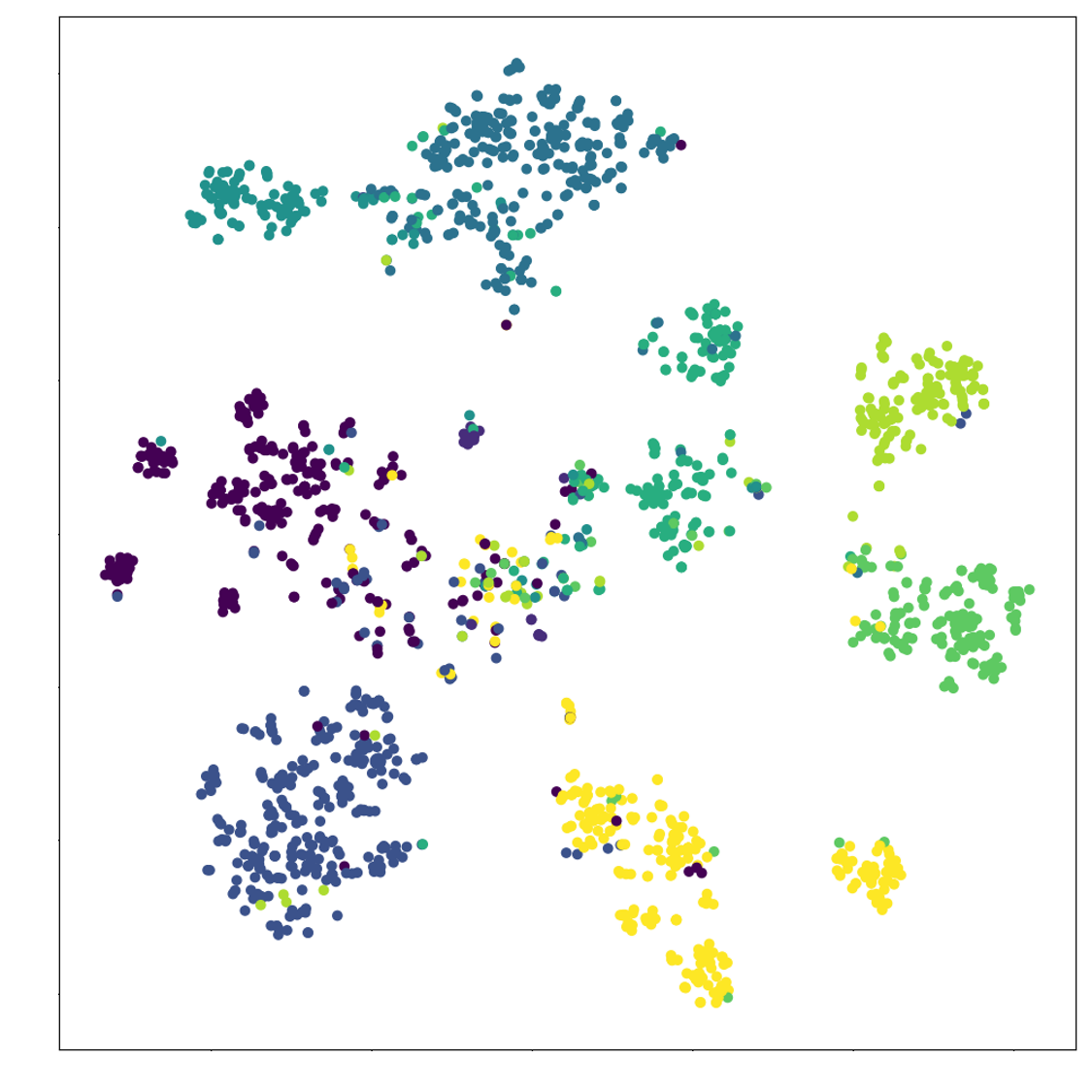}\\
\end{tabular}
\caption{Visualization of Multiwoz 2.2 test dataset. The left is utterance representation before training, and the right is after training.}
\label{fig:training}
\end{figure}

Recently, there have been several successful studies using contrastive learning (CL) in the computer vision and language domain \cite{chen2020simple, liu2020hybrid, wu2018unsupervised, gunel2020supervised} and CL shows more generalize and robustness to language model training than cross-entropy loss \cite{gunel2020supervised}. Following their success, we utilize supervised contrastive learning (SCL) \cite{khosla2020supervised} in fine-tuning.

SCL is a modified version of the CL approach, which utilize the label information. In CL, only the anchor and its augmented object are regarded as positive objects, and others are used as negative object in training. However, SCL set the same label objects as positive and others as negative, so more accurate embedding representation learning is possible.

For the $N$ randomly sampled object ${\{x_k, y_k\}_{k=1...N}}$, mini batch used for training is consists of $2N$ pair,${\{\tilde{x}_k, \tilde{y}_k\}_{k=1...2N}}$. In ${\{\tilde{x}_k, \tilde{y}_k\}_{k=1...2N}}$, ${\{\tilde{x}_k, \tilde{y}_k\}_{k=1...N}}$ is same as original sampled ${\{x_k, y_k\}_{k=1...N}}$ and ${\{\tilde{x}_k, \tilde{y}_k\}_{k=N...2N}}$ is augmented pair of original sampled data. Word-net \cite{miller1995wordnet} based synonym augmentation\footnote{https://github.com/makcedward/nlpaug} is used for augmentation. $\Phi$ denotes the SBERT encoder and $\tau$ is scalar temperature parameter to adjust the separation strength. The overall loss is given by the following equations, and visualization of the training results are shown in Figure~\ref{fig:training}:

\begin{equation}
\label{eq:1}
\small
    \mathcal{L}^{sup}=\sum_{i=1}^{2N}\mathcal{L}_{i}^{sup}
\end{equation}

\begin{equation}
\label{eq:2}
\small
\begin{split}
{L}_{i}^{sup}={} & -\frac{1}{2N_{\tilde{y_i}}-1} \sum_{j=1}^{2N} \mathbf{1}_{i\ne j}\cdot\mathbf{1}_{\tilde{y_i}=\tilde{y_j}}\cdot\\    
    &  \log \frac{\exp(\Phi_i\cdot \Phi_j/\tau)}{\sum\nolimits_{k=1}^{2N}\mathbf{1}_{i\ne k}\exp (\Phi_i\cdot \Phi_j/\tau)}
\end{split}
\end{equation}

\subsection{Intent label generation}
\label{label_generation}
To improve the explainability of the clustering results, we automatically generated the semantic labels from the clustering results. Following the intent datasets, which usually represent the intent name as a verb and object pair \cite{zang2020multiwoz, coucke2018snips, rastogi2020schema}, we also named our induced clusters as Verb-Object forms using a dependency parser. In the previous method, \citet{liu2021open} counted the common Verb-Object pairs in the cluster and used the most common pair as the cluster name. However, this method did not create a proper label when detailed words appeared in the object. For example, a pair of \code{call-son} and \code{call-daughter} cannot be grouped as \code{call-child} using the previous method. 

To overcome this limitation, we propose a method that uses the object word’s hypernyms. By adopting hypernyms, we could obtain a proper word containing detailed information. More precisely, we generate \textit{verb-hyper(object)} and \textit{verb-hyper(hyper(object))} pairs from existing Verb-Object pairs and calculate the most common pair from this generated result. We employ this rule when the number of the most common pair and second place does not differ by more than $\alpha$ times, and we set $\alpha$ to 2.0 in the experiment. Word-net \cite{miller1995wordnet} is used to get the hypernyms.
\section{Experiment}

\begin{figure*}
\centering
\begin{tabular}{cccc}
\hline
Dev (Insurance) & Test1 (Banking) &  Test2 (Finance) & SGD (Tourism)\\
\hline
\multicolumn{4}{c}{\textit{Baseline}} \\ \hline
\includegraphics[width=0.2\textwidth]{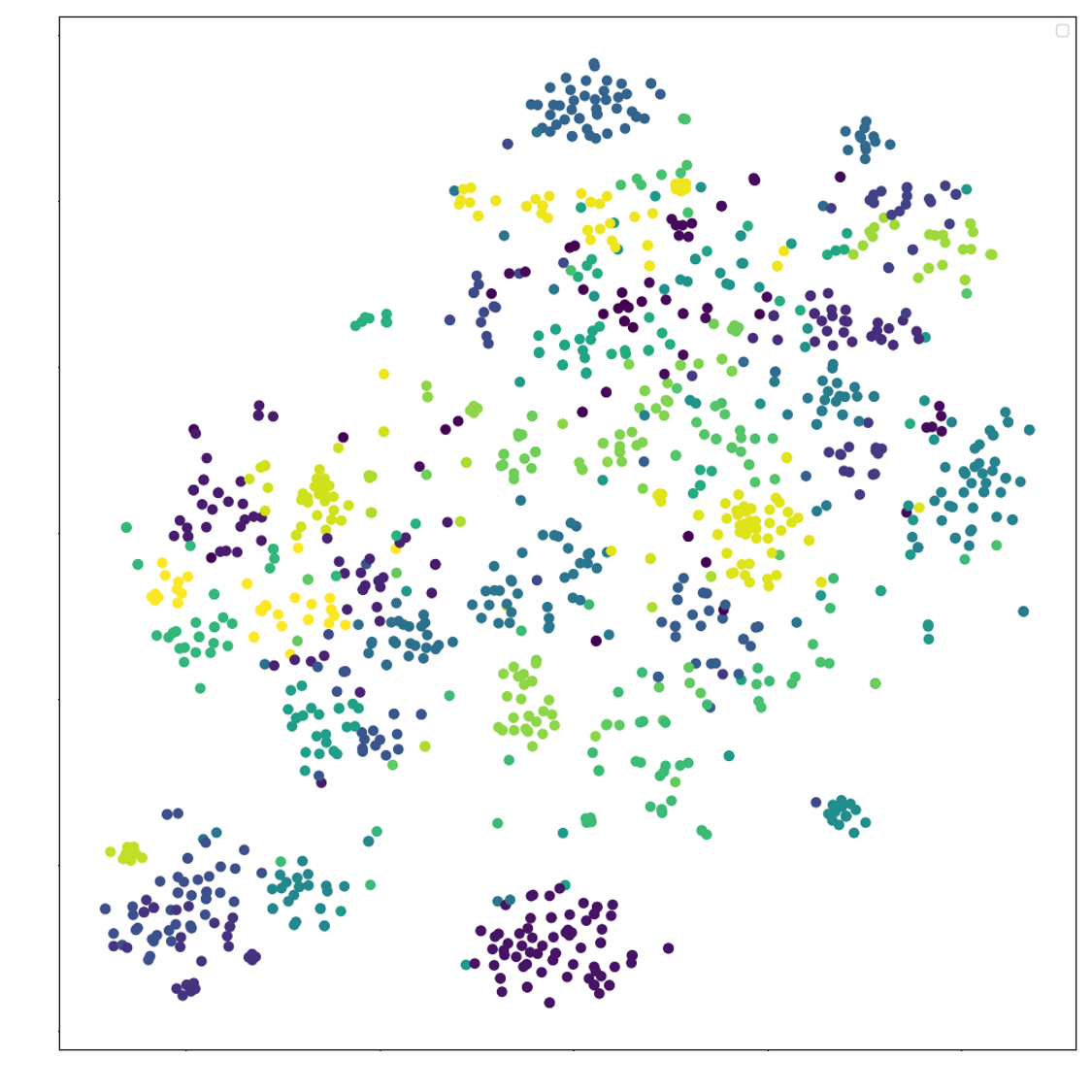}& 
\includegraphics[width=0.2\textwidth]{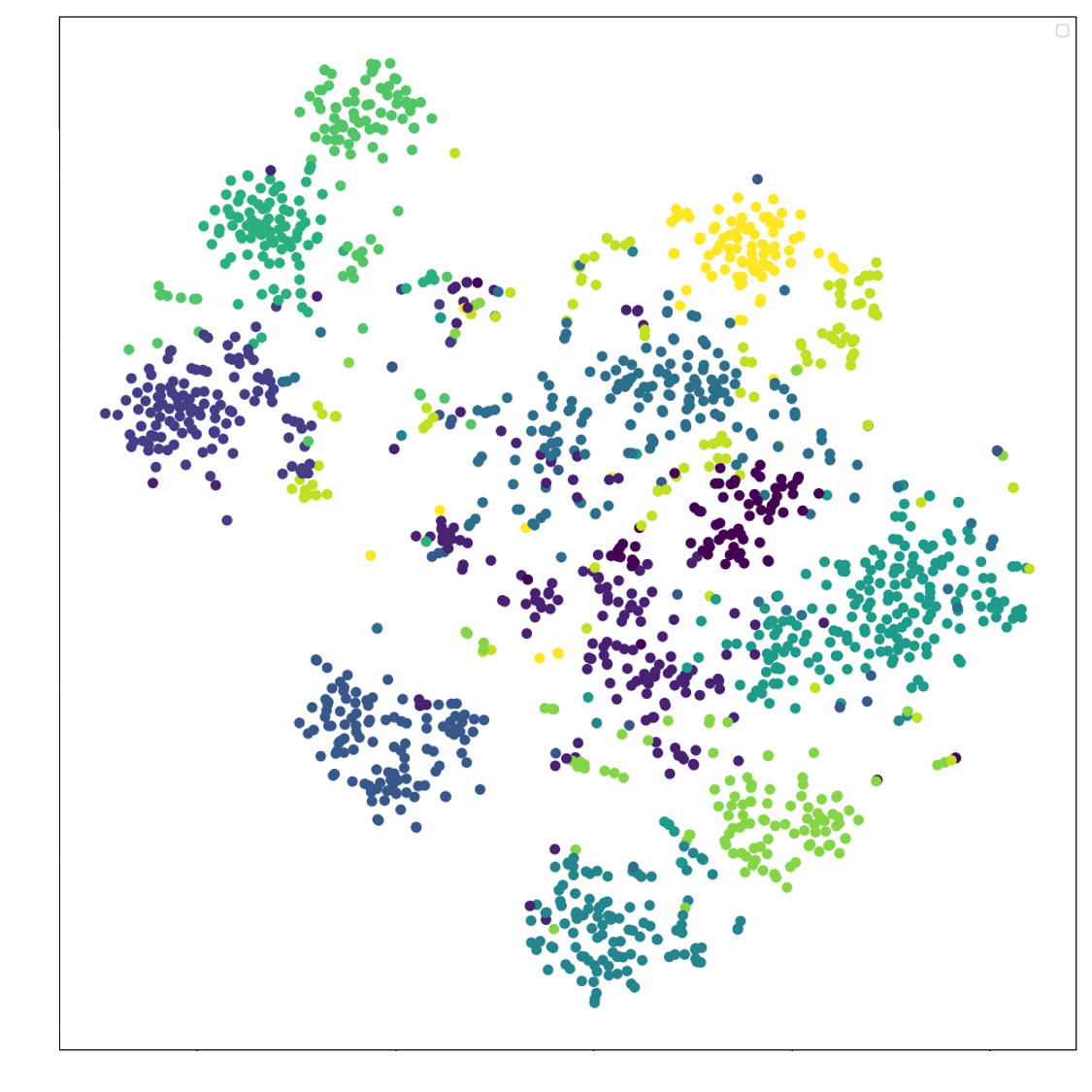}&
\includegraphics[width=0.2\textwidth]{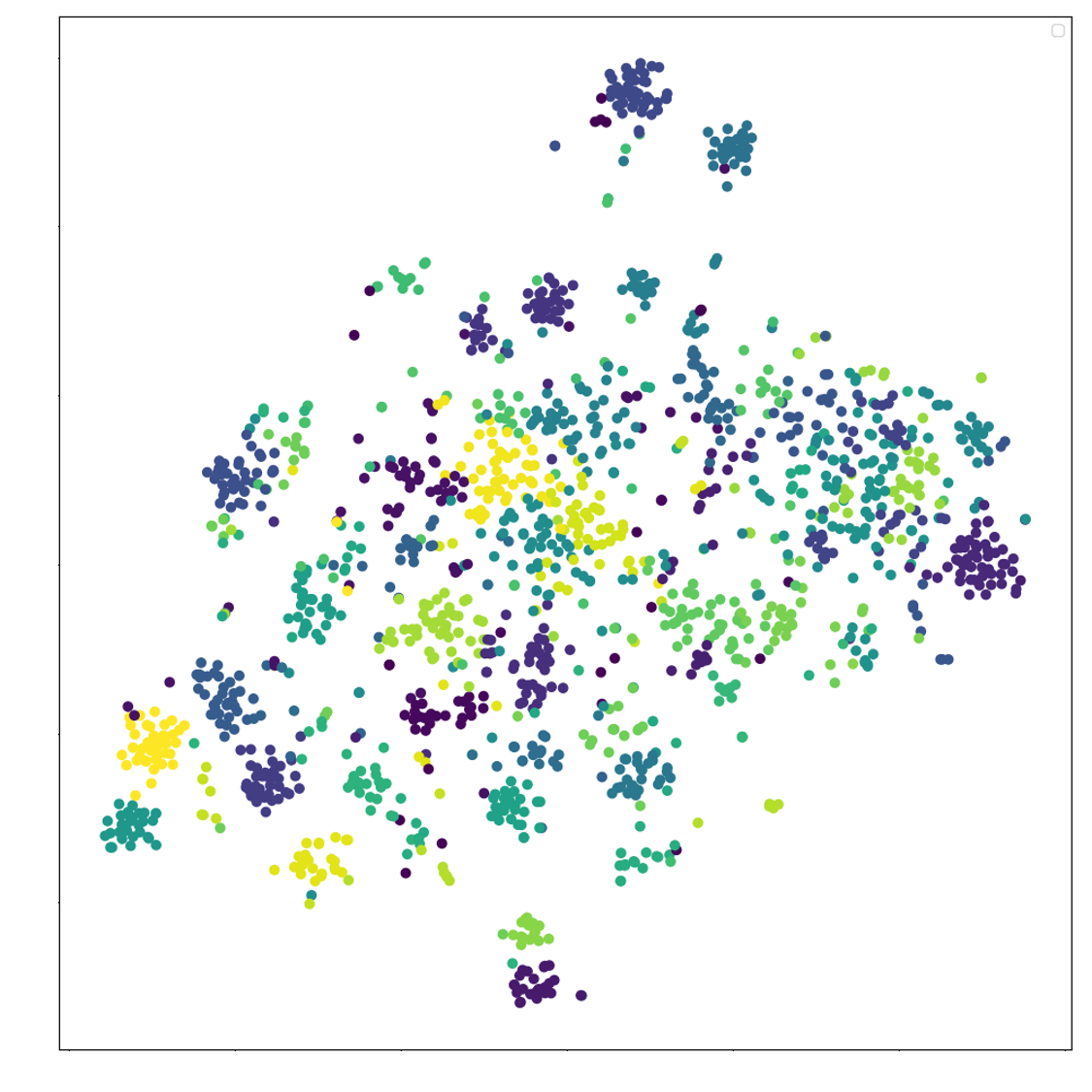}&
\includegraphics[width=0.2\textwidth]{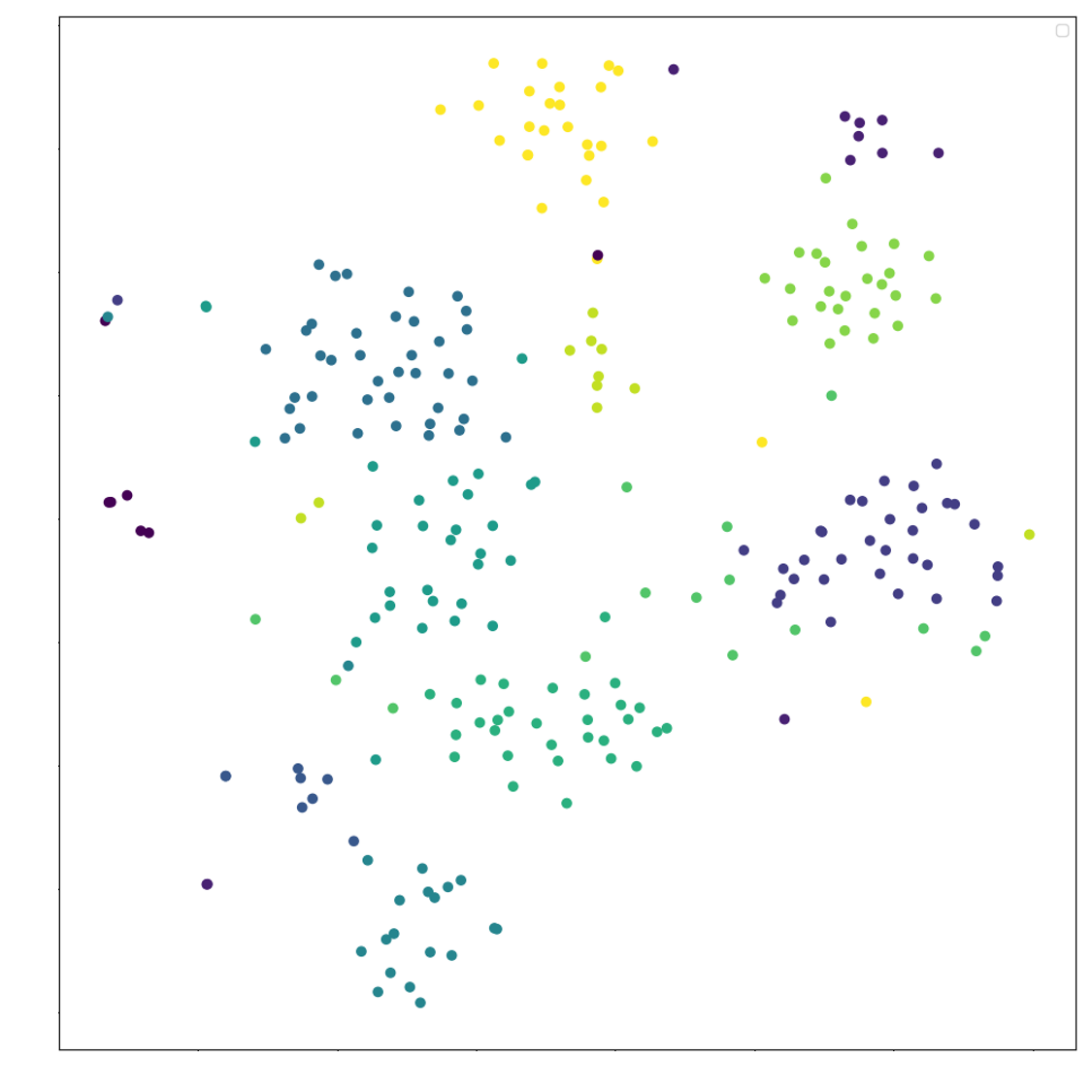}\\
\hline
\multicolumn{4}{c}{\textit{DORIC}} \\ \hline
\includegraphics[width=0.2\textwidth]{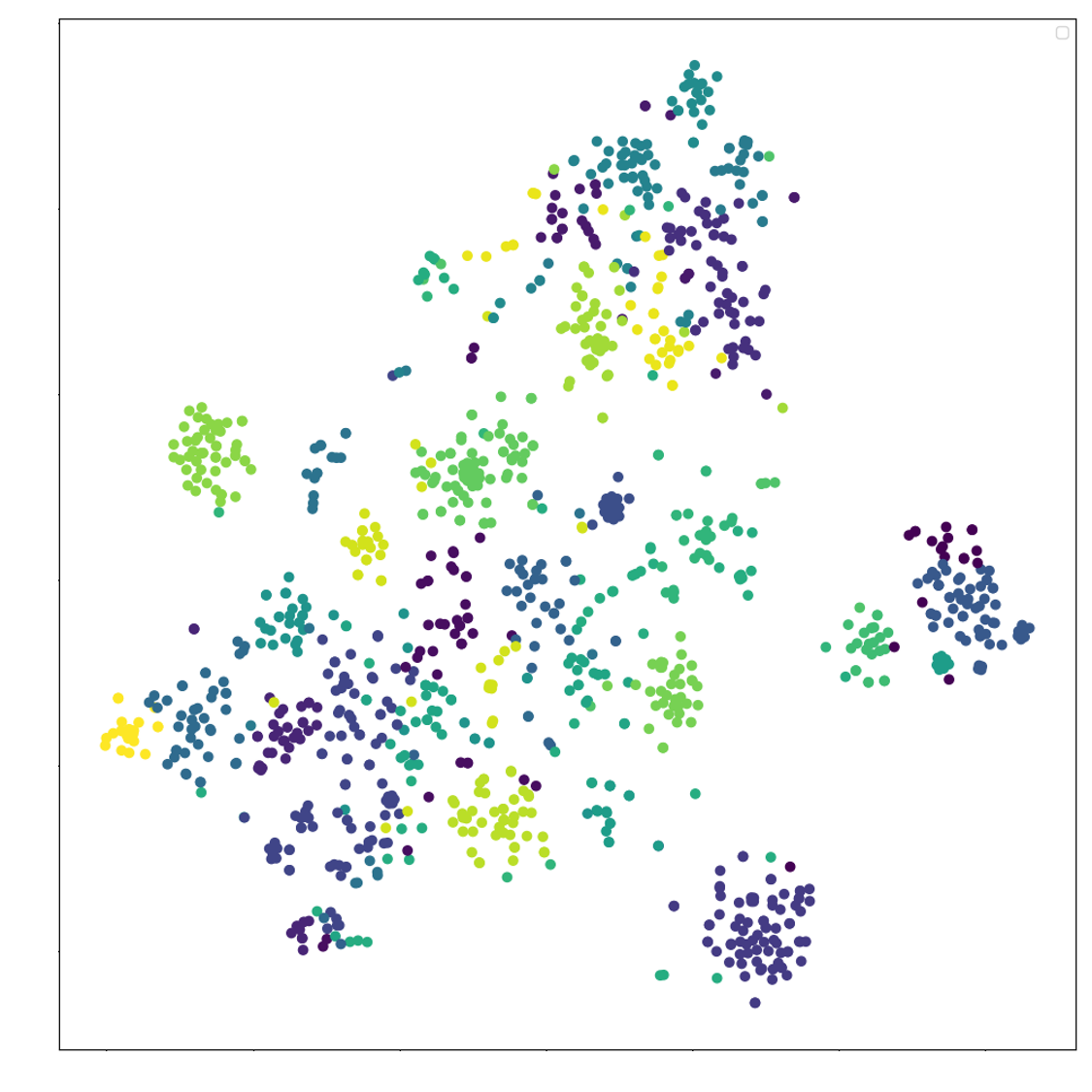}&
\includegraphics[width=0.2\textwidth]{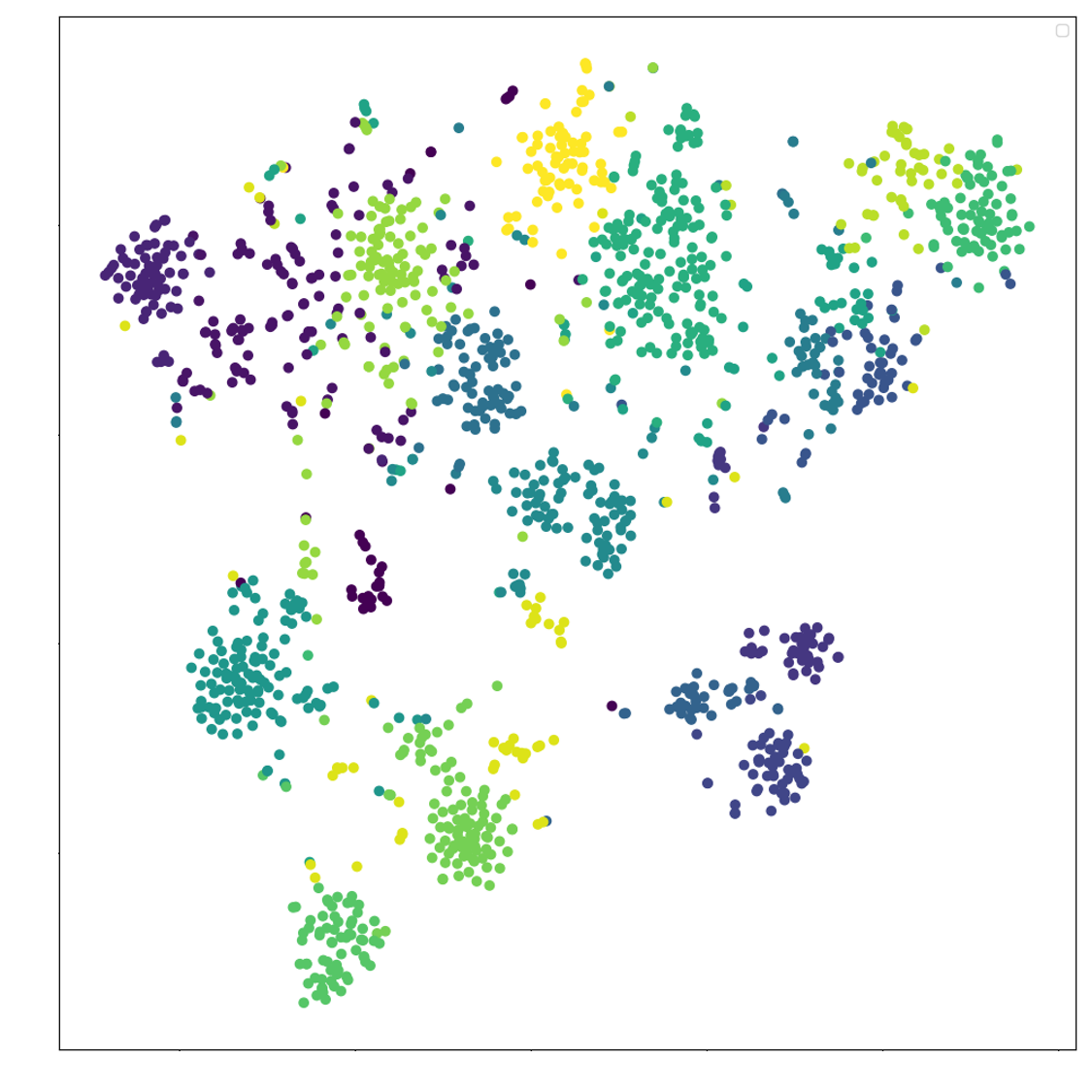}&
\includegraphics[width=0.2\textwidth]{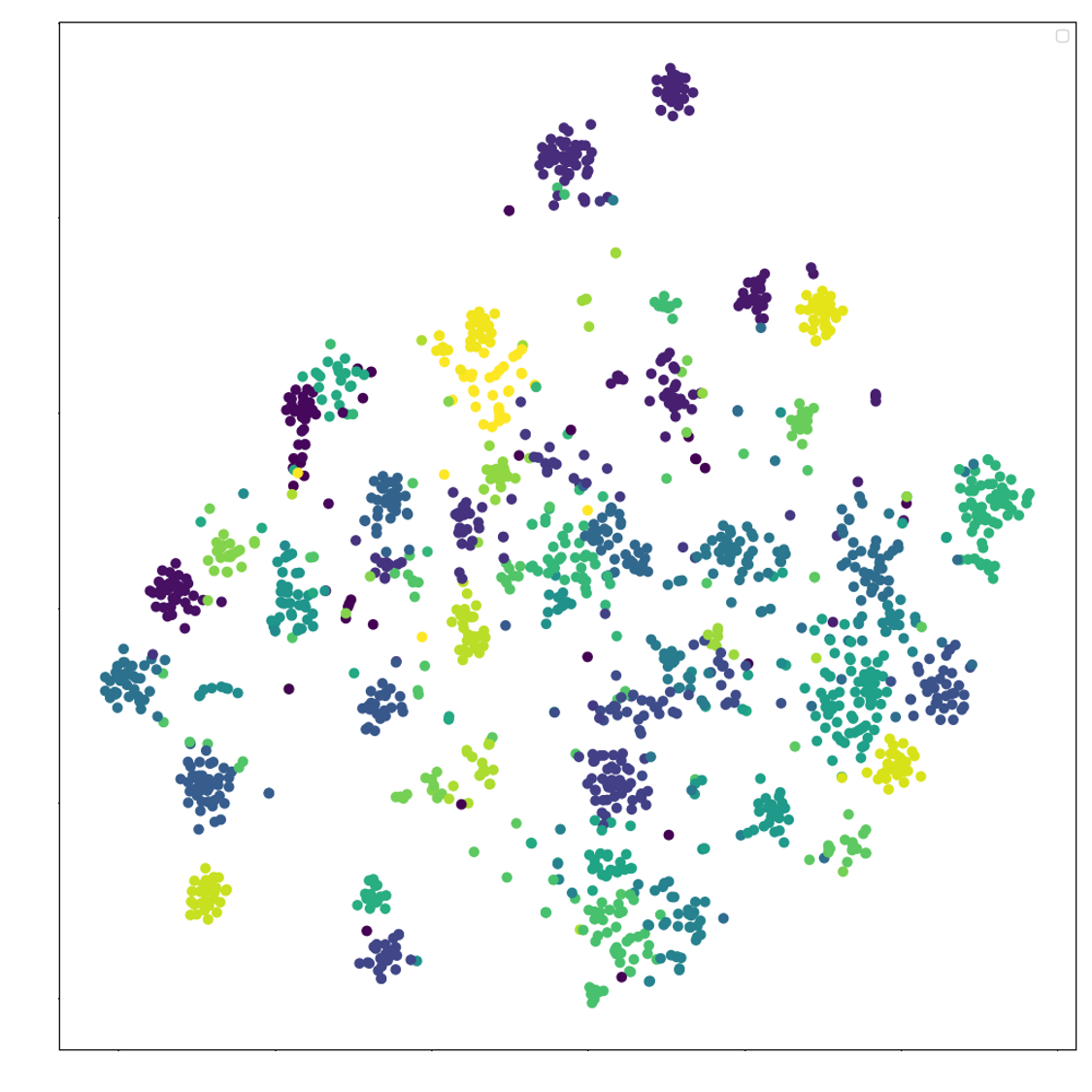}&
\includegraphics[width=0.2\textwidth]{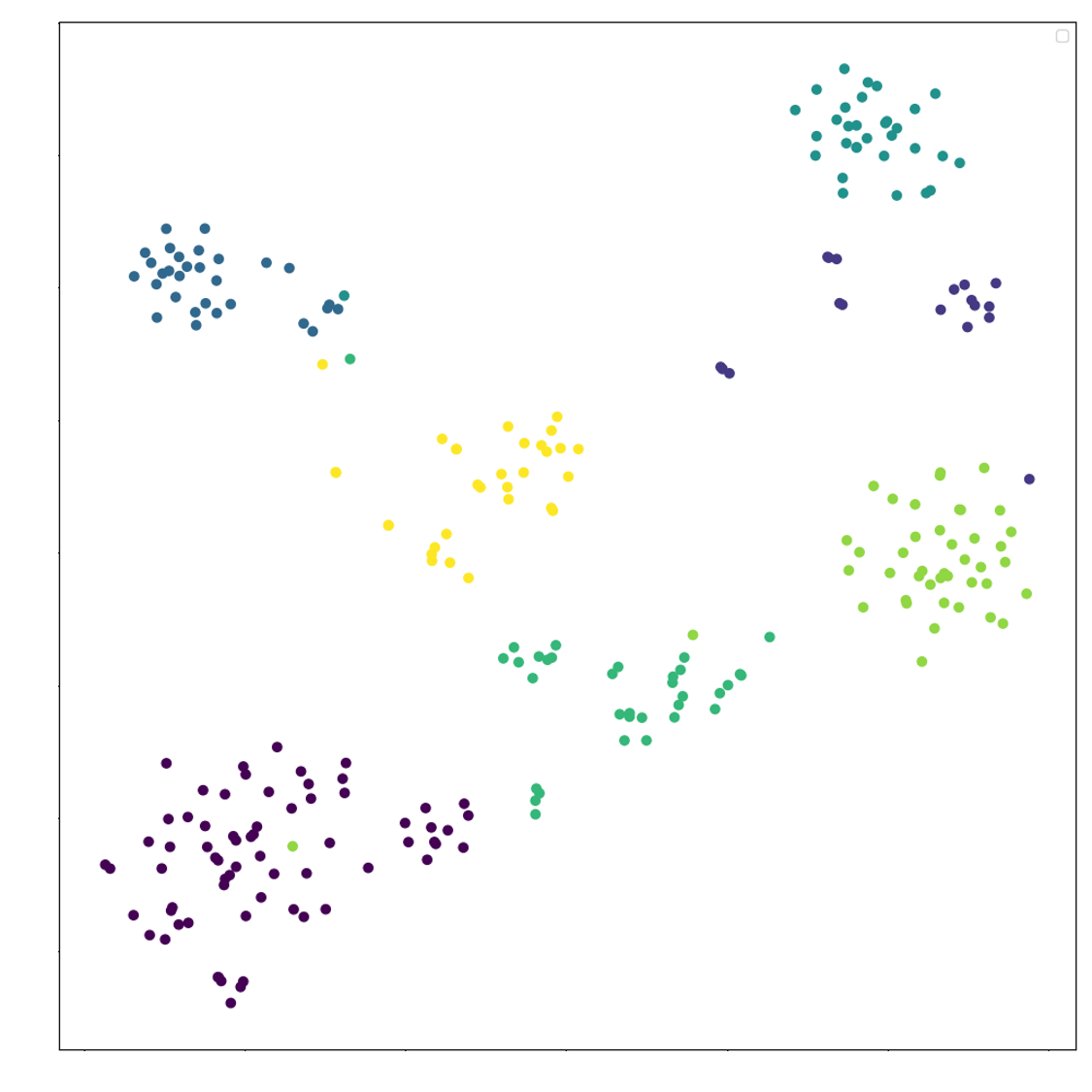}\\
\hline

\end{tabular}
\caption{Visulization on dev (insurance), test1 (banking), test2 (finance) and SGD (tourism) dataset.}
\label{fig:base_ours}
\end{figure*}

\subsection{Experimental setup}
\textbf{Dataset}
\noindent
To demonstrate the performance of our model, we used the development (dev) (insurance), test1 (banking), and test2 (finance) datasets. These datasets have domains that are different from the fine-tuning dataset (tourism), so we were able to examine our method's effectiveness in diverse domains. Additionally, we used the Schema-Guided Dialogue Dataset (SGD) dataset; we extracted tourism-related domains from the SGD dataset to make the same domain environment with fine-tuning dataset. The number of intents for each dataset is shown in Table~\ref{tab:dataset}.

\begin{table}[h]
\centering
\small
\begin{tabular}{lll}
\hline
\textbf{Dataset} & \textbf{Domain} & \textbf{\# of intents}\\ \hline
Dev  & Insurance & 22\\
Test1 & Banking & 29\\
Test2 & Fianace & 39\\
SGD & Tourism & 6\\

\hline

\end{tabular}
\caption{Domain and number of intents type for each dataset.}
\label{tab:dataset}
\end{table}

\noindent
\textbf{Evaluation} 
NMI and accuracy were the primary metrics used for the evaluation, and to provide additional metrics, precision was also used. The higher NMI value denotes that clustering has reduced more entropy. 1:1 alignments between the induced intents and the gold intents were computed by the Hungarian algorithm \cite{kuhn1955hungarian}.\\

\noindent\textbf{Setup} 
We employ the pre-trained SBERT \cite{reimers2019sentence} for the baseline embedding model. The pre-trained parameters were from the huggingface \cite{wolf2019huggingface} \textit{all-mpnet-base-v2} version. In the SCL function, we set the $\tau$ as 0.07 and trained a maximum of five epochs with early stopping. In the K-means clustering, we set the minimum cluster number as five and the max cluster number as 50 and use silhouette score for comparing the clustered result, which is based on tightness and separation \cite{rousseeuw1987silhouettes}.

\subsection{Intent clustering result}

\begin{table}[h]
\centering
\small

\begin{tabular}{ll|lll}

\hline
\textbf{Model} & \textbf{Data} & \textbf{NMI}& \textbf{ACC} & {\fontsize{8}{7}\textbf{Precision}} \\ \hline

\multicolumn{5}{c}{\textit{Different Domain with Fine-tuning Dataset}} \\ \hline
Baseline & Dev(Insurance)                & 59.31 & 46.14 &65.98 \\
DORIC   & Dev(Insurance)                 & \textbf{65.16} & \textbf{56.68}  &\textbf{67.63}\\
\hline
Baseline & Test1(Banking) & 65.71 & 51.85   &60.68\\
DORIC   & Test1(Banking)   & \textbf{71.02} & \textbf{52.35}  & \textbf{73.92}\\
  \hline
Baseline & Test2(Finance)         & 60.26 & 59.75  &69.25 \\
DORIC   &  Test2(Finance)           & \textbf{69.64} & \textbf{65.14} &\textbf{75.14} \\
 \hline
\multicolumn{5}{c}{\textit{Same Domain with Fine-tuning Dataset}} \\ \hline
Baseline & SGD(Tourism)  & 60.54 & 63.67  &49.90\\
DORIC   & SGD(Tourism) & \textbf{65.32} & \textbf{68.36}  &\textbf{51.00}\\

\hline

\end{tabular}

\caption{Comparison of baseline and DORIC in different dataset. NMI, ACC and Precision are reported.}
\label{tab:main_result}
\end{table}
The results of DORIC after evaluation on the dev (insurance), test1 (finance), test2 (banking), and SGD (tourism) datasets are shown in Table~\ref{tab:main_result}. These results show that our model outperforms the baseline model in terms of the NMI, ACC, and precision on all datasets. Except for the SGD dataset, the dataset’s domains are all different from the fine-tuning dataset MultiWOZ2.2 (tourism), which demonstrates that our intent induction framework is robust to diverse domain datasets. The visualization of experimental results in Figure~\ref{fig:base_ours} also exhibits the aligned result with Table~\ref{tab:main_result}; compared to the baseline, DORIC embeds utterances with the same label at a closer distance.

\subsection{Intent label generation with hypernym}
\begin{table}[]
\centering
\small
{
\begin{tabular}{lll}
\hline

\textbf{Idx} & \textbf{Generated name}& \textbf{Ground-truth}    \\ \hline
0               & create-account  & CreateAccount             \\
1 & cancel-billing & CancelAutomaticBilling    \\
2\dag & \textbf{add-child} & AddDependent \\
3 & report-accident & ReportAutomobileAccident\\
4 & change-address & ChangeAddress\\
5\dag & get-quote & GetQuote\\
6 & change-plan & ChangePlan\\
7 & file-claim & FileClaim \\
8  & pay-bill & PayBill\\
9 & check-balance & CheckAccountBalance\\
10 & change-question & ChangeSecurityQuestion\\
11 & cancel-plan & CancelPlan\\
\hline
\multicolumn{3}{c}{\textit{Without hypernym}} \\ 
\hline
2 & \textbf{add-son} & AddDependent \\
5 & get-quote & GetQuote\\
\hline
\end{tabular}%
}
\caption{Example of generated intent labels and ground truth. Cluster name with \dag means using hypernym.}
\label{tab:label_generation}
\end{table} 

Table~\ref{tab:label_generation} shows examples of the generated intent labels, and cluster with \dag denotes the clusters with the hypernyms following section~\ref{label_generation}. As shown in the table, our proposed method successfully explains the cluster results compared to the ground truth label. Furthermore, using hypernyms enables the grouping of detailed information in the cluster. For instance, Cluster 2 obtains a more comprehensive label, \code{add-child} than \code{add-son} by using the hypernyms. We also present the more detailed results for the dev and test data in Appendix A.1.

\section{Analysis}
\subsection{Verb-Object structure in fine-tuning}
\begin{table}[h!]
\centering
\small
\begin{tabular}{ll|ll}
\hline
\textbf{Method} & \textbf{Dataset (domain)} & \textbf{NMI}& \textbf{ACC} \\ \hline

\multicolumn{4}{c}{\textit{Different Domain with Fine-tuning Dataset}} \\ \hline

Sentence & Dev (Insurance)  & 62.13 & 55.35 \\
Verb-Obj    & Dev (Insurance)          & \textbf{65.16} & \textbf{56.68} \\
\hline
Sentence & Test1 (Banking)     & 68.91 & \textbf{53.22} \\
Verb-Obj     & Test1 (Banking)      & \textbf{71.02} & 52.35 \\
 \hline

Sentence & Test2 (Finance)              & 64.94 & \textbf{67.07} \\
Verb-Obj     & Test2 (Finance)  & \textbf{69.64} & 65.14 \\
 \hline

\multicolumn{4}{c}{\textit{Same Domain with Fine-tuning Dataset}} \\ \hline
Sentence & SGD (Tourism)  & 65.24 & 68.35 \\
Verb-Obj     & SGD (Tourism) & \textbf{65.32} & \textbf{68.36} \\
\hline
\end{tabular}
\caption{The NMI and accuracy result on DSTC11 development, test, and SGD dataset according to fine-tuning utterance format.}
\label{tab:extract}
\end{table}

To examine the effect of extracting Verb-Object structures from the sentence, we compare our proposed method with methods that use the whole sentence during the fine-tuning stage (Table~\ref{tab:extract}). Using the Verb-Object structure demonstrates superior NMI results in both different-domain and same-domain environments; this result indicates that fine-tuning with Verb-Object information has helped reduce the clustering uncertainty. However, the accuracy doesn't significantly differ between the Verb-Object form and the whole sentence form in the tourism domain, which is identical to the fine-tuning dataset domain.

\subsection{Analysis of loss}

\begin{table}[h]
\centering
\small
\begin{tabular}{ll|ll}
\hline
\textbf{Loss function} & \textbf{Dataset (domain)} & \textbf{NMI}& \textbf{ACC} \\ \hline

\multicolumn{4}{c}{\textit{Different Domain with Fine-tuning Dataset}} \\ \hline
Cross Entropy & Dev (Insurance)  & 61.98 & 53.69 \\
SCL    & Dev (Insurance)          & \textbf{65.16} & \textbf{56.68} \\
\hline
Cross Entropy & Test1 (Banking)              & 67.67 & 52.16 \\
SCL   & Test1 (Banking)    & \textbf{71.02} & \textbf{52.35} \\
\hline
Cross Entropy & Test2 (Finance)     & 64.09 & 63.07 \\
SCL   & Test2 (Finance)    & \textbf{69.64} & \textbf{65.14} \\
\hline

\multicolumn{4}{c}{\textit{Same Domain with Fine-tuning Dataset}} \\ \hline
Cross Entropy & SGD (Tourism) & 64.11 & \textbf{70.31} \\
SCL   & SGD (Tourism) & \textbf{65.32} & 68.36 \\
\hline

\end{tabular}
\caption{The NMI and accuracy result on DSTC11 development, test, and SGD dataset according to the loss function.}
\label{tab:class_contrast}
\end{table}

To investigate the effect of SCL during fine-tuning, we compare the result with the cross-entropy loss in Table~\ref{tab:class_contrast}. In most cases, the SCL loss demonstrates better results by a large margin; however, on the SGD dataset, the NMI and ACC results were slightly or no different than the cross-entropy loss. Considering that the SGD dataset is the only dataset with the same domain with the fine-tuning dataset (tourism), this result indicates that SCL is more useful when it is used in a domain-across environment.

\section{Conclusion}
In this paper, we describe our solution for the DSTC11 intent induction competition. We leveraged the SBERT model to embed sentences and fine-tuned the model using dependency parsing results. Additionally, we used supervised contrastive loss during fine-tuning to make the model robust in multiple domains. During the analysis, both dependency parsing and SCL helped to make the intent induction model more domain robust. Furthermore, our intent label generation with hypernym methods allows us to explain the clustering results. According to the results, our approach achieved 3rd place in terms of the precision score and demonstrated better NMI and accuracy compared to the baseline model.

\section*{Limitations}
Our contribution has two limitations. First, although DORIC shows superior performance in the domain across the environment, the increase was insignificant in the same domain environment. Second, we thoroughly examine the embedding methods, but we adapt this method only to the K-means clustering. In the future, we plan to devise a progressed clustering method that fits our embedding method.

\section*{Acknowledgements}

This work was partly supported by Institute of Information \& communications Technology Planning \& Evaluation (IITP) grant funded by the Korea government(MSIT) (No.2019-0-01906, Artificial Intelligence Graduate School Program(POSTECH)) and  ITRC(Information Technology Research Center) support program(IITP-2023-2020-0-01789).

\bibliography{anthology,custom,jihyun}
\bibliographystyle{acl_natbib}

\appendix
\label{sec:appendix}
\onecolumn  
\section{Appendix}
\subsection{Generated cluster label name in detail.}
\begin{table*}[h!]
\centering
\tiny
{
\begin{tabular}{llll}
\hline

\textbf{Cluster} & \textbf{TOP 3 Verb-Object Pairs}      & \textbf{Example}        & \textbf{Ground-truth}                                   \\ \hline
0               & ('create-account', 13), ('make-account', 5), ('open-account', 3)         & $\sim$create an account for my new renter policy.           & CreateAccount             \\
1                & ('cancel-billing', 6), ('stop-payment', 3), ('cancel-payment', 2)        & $\sim$to cancel the automatic billing of my account $\sim$      & CancelAutomaticBilling   \\
2\dag                 & ('add-child', 10), ('add-son', 5), ('add-male\_offspring', 5)            & Oh good, help me add my son.                                     & AddDependent\\
3                & ('report-accident', 10), ('file-claim', 1), ('start-process', 1)         & Hey yeah, I have to call and report an accident.            & ReportAutomobileAccident  \\
4               & ('change-address', 31), ('update-address', 9), ('update-information', 2) & Hi, I'd like to change my address to a new one $\sim$        & ChangeAddress\\
5\dag                & ('get-quote', 10), ('get-punctuation', 10), ('get-interruption', 10)     & $\sim$to get a quote from you guys.                                 & GetQuote\\
6               & ('change-plan', 9), ('upgrade-plan', 1), ('get-quote', 1)                & Eee yeah I think I want to change my plan                       & ChangePlan \\
7               & ('file-claim', 20), ('report-claim', 3), ('make-claim', 2)               & I'd like to file a property claim.                                & FileClaim\\
8               & ('pay-bill', 17), ('get-bill', 2), ('take-care', 2)                      & Well, I'm calling to pay my bill.                                   & PayBill\\
9               & ('check-balance', 4), ('pay-bill', 2), ('confirm-balance', 2)            & Yes I need to check the balance $\sim$                       & CheckAccountBalance\\
10               & ('change-question', 10), ('remember-question', 2), ('keep-stuff', 1)     & $\sim$to change my security question and answer.          & ChangeSecurityQuestion \\
11               & ('cancel-plan', 24), ('cancel-policy', 8), ('cancel-insurance', 6)       & $\sim$I need to cancel my plan.                            & CancelPlan\\ \hline
\end{tabular}%
}
\caption{Generated intent cluster label, example, and ground truth on dev (insurance) dataset. Cluster name with \dag means using hypernym.}
\label{dev:intent label induction}
\end{table*}

\begin{table}[h!]
\centering
\resizebox{\columnwidth}{!}{%
\begin{tabular}{llll}
\hline
\textbf{Cluster} & \textbf{TOP 3 Verb-Object Pairs}                  & \textbf{Example}&  \textbf{Ground-truth} \\ \hline
0                & ('dispute-transaction', 17), ('have-transaction', 3), ('dispute-charge', 2)            & almost. I need to dispute this transaction I found for Piggly Wiggly.       & DisputeCharge         \\
1                & ('open-account', 56), ('set-banking', 2), ('set-account', 2)                           & $\sim$ so I would need to open a checking account then, right?                         & OpenBankingAccount    \\
2                & ('update-address', 19), ('change-address', 7), ('do-address', 1)                       & I'd like to update my address I believe when $\sim$                                  & UpdateStreetAddress   \\
3\dag            & ('transfer-money', 23), ('transfer-medium\_of\_exchange', 23), ('transfer-standard', 18) & And transfer the money from that to my checking.                               & InternalFundsTransfer \\
4                & ('check-balance', 68), ('get-balance', 8), ('give-balance', 7)                         & yes I need to check the my account balance $\sim$                                      & CheckAccountBalance   \\
5\dag            & ('make-transfer', 23), ('make-movement', 23), ('make-change', 23)                      & Yeah, I needed to make a wire transfer $\sim$                                          & ExternalWireTransfer  \\
6                & ('find-branch', 19), ('locate-branch', 5), ('help-branch', 4)                          & Yes. I need help with the with finding the nearest branch to my location.          & FindBranch  \\ 
\hline
\end{tabular}
}
\caption{Generated intent cluster label, example, and ground truth on test1 (banking) dataset. Cluster name with \dag means using hypernym.}
\label{banking: intent label induction}
\end{table}

\begin{table}[h!]
\centering
\resizebox{\textwidth}{!}{%
\begin{tabular}{llll}
\hline
\textbf{Cluster} & \textbf{TOP 3 Verb-Object Pairs}                                    & \textbf{Example}                                        & \textbf{Ground-truth} \\ \hline
0                & ('check-balance', 18), ('check-account', 4), ('check-checking', 2)  & $\sim$ I am calling to check my account balance.                  & CheckAccountBalance   \\
1\dag            & ('update-number', 13), ('update-amount', 13), ('update-assets', 13) & $\sim$ and next update my phone number for my business$\sim$           & UpdatePhoneNumber     \\
2                & ('check-balance', 16), ('get-balance', 6), ('know-balance', 3)      & I need to check the balance on my credit card please         & CheckAccountBalance   \\
3                & ('update-address', 17), ('change-address', 6), ('change-piece', 1)  & yeah I need to update my street address on $\sim$                  & UpdateStreetAddress   \\
4\dag            & ('open-account', 11), ('open-record', 11), ('open-evidence', 11)    & $\sim$ I was thinking about opening another account for $\sim$   & OpenAccount \\
5                & ('add-user', 10), ('have-access', 2), ('bring-people', 1)           & Yeah I need to add some additional users to my account          & AddUserToAccount      \\
6\dag            & ('make-transfer', 13), ('make-movement', 13), ('make-change', 13)   & Hi Jerry. I need to make a wire transfer man.           & MakeTransfer      \\
\hline
\end{tabular}%
}
\caption{Generated intent cluster label, example, and ground truth on test2 (finance) dataset. Cluster name with \dag means using hypernym.}
\label{finance: intent label induction}
\end{table}

\end{document}